%% file: main.tex
\documentclass[conference]{IEEEtran}
\IEEEoverridecommandlockouts
\usepackage{cite}
\usepackage{amsmath,amssymb,amsfonts}
\usepackage[ruled,linesnumbered]{algorithm2e}
\usepackage{algpseudocode}
\usepackage{commath}
\usepackage{tabularx,booktabs}
\usepackage{tikz}
\usepackage{graphicx}
\usepackage{textcomp}
\usepackage{xcolor}
\usepackage{booktabs}
\usepackage{multirow}
\usepackage{multicol}
\usepackage{array}
\usepackage{xspace}
\usepackage{bm}
\usepackage{caption}
\usepackage{subcaption}
\usepackage{makecell}
\usepackage{hyperref}

\newcommand{\nickname}{\texttt{EMO}\xspace}
\newcolumntype{P}[1]{>{\centering\arraybackslash}p{#1}}

\begin{document}
\title{\nickname: Edge Model Overlays to Scale Model Size in Federated Learning}

\author{
    Di~Wu,
    Weibo~He,
    Wanglei~Feng,
    Zhenyu~Wen,
    Bin~Qian,
    and~Blesson~Varghese
    \IEEEcompsocitemizethanks{
        \IEEEcompsocthanksitem Corresponding author: Zhenyu Wen.
        \IEEEcompsocthanksitem Di Wu and Blesson Varghese are with the School of Computer Science, University of St Andrews, UK. 
        \IEEEcompsocthanksitem Weibo He and Wanglei Feng are with the Institute of Cyberspace Security and College of Information Engineering, Zhejiang University of Technology, China. 
        \IEEEcompsocthanksitem Zhenyu Wen is with the Institute of Cyberspace Security and College of Information Engineering, Zhejiang University of Technology, China; University of Science and Technology of China, China.
        \IEEEcompsocthanksitem Bin Qian is with the State Key Laboratory of Industrial Control Technology, Zhejiang University, China.

    }
}

\maketitle
\input{abstract}
\input{introduction-v2}
\input{related_work}

\input{challenges}
\input{elo-v2}
\input{evaluation-v2}
\input{conclusion}

\section*{Acknowledgment}
This work was supported by UK Research and Innovation grant EP/Y028813/1; National Nature Science Foundation of China under Grant 62472387, China; Postdoctoral Science Foundation under Grant 2023M743403 and Zhejiang Provincial Natural Science Foundation of Major Program (Youth Original Project) under Grant LDQ24F020001.

\bibliography{main}
\bibliographystyle{plain}
\end{document}

%% file: abstract.tex
\begin{abstract}
Federated Learning (FL) trains machine learning models on edge devices with distributed data. However, the computational and memory limitations of these devices restrict the training of large models using FL. Split Federated Learning (SFL) addresses this challenge by distributing the model across the device and server, but it introduces a tightly coupled data flow, leading to computational bottlenecks and high communication costs. We propose \nickname as a solution to enable the training of large models in FL while mitigating the challenges of SFL. \nickname introduces Edge Model Overlay(s) between the device and server, enabling the creation of a larger ensemble model without modifying the FL workflow. The key innovation in \nickname is Augmented Federated Learning (AFL), which builds an ensemble model by connecting the original (smaller) FL model with model(s) trained in the overlay(s) to facilitate horizontal or vertical scaling. This is accomplished through three key modules: a hierarchical activation replay cache to decouple AFL from FL, a convergence-aware communication controller to optimize communication overhead, and an ensemble inference module. Evaluations on a real-world prototype show that \nickname improves accuracy by up to 17.77\% compared to FL, and reduces communication costs by up to 7.17$\times$ and decreases training time by up to 6.9$\times$ compared to SFL.
\end{abstract}

%% file: introduction-v2.tex
\section{Introduction}

Machine Learning (ML) applications in 
language and vision tasks usually use centralized data for training models~\cite{kaplan2020scaling}. 
However, distributed and private data generated or stored on devices at the edge of the network, such as on mobile phones, is expected to play an increasingly important role in training~\cite{deng2020edge}. 
Federated Learning (FL) is a proposed privacy-aware solution for training models using distributed data.

However, the \textit{models produced by existing FL solutions are limited in several ways}. Firstly, the size of the model is bound by hard constraints, such as the memory of the device participating in FL~\cite{saikumar2024neuroflux}. Secondly, the relatively limited computing capability on devices results in impractical training times~\cite{wu2022fedadapt,zhang2022federated}. For instance, devices used in FL, such as Nvidia Nano, have 8 to 50 times less memory and 10 to 1,000 times fewer FLOPS than resources available in centralized servers, such as Nvidia A100 GPU~\cite{pfeiffer2023federated}. 

Therefore, \textit{the models trained using edge devices in classic FL systems are small and have limited learning performance.} 

Split Federated Learning (SFL) has been proposed to mitigate the above shortcoming by offloading certain layers of the model from the device to a server~\cite{thapa2022splitfed}.
This reduces the memory use and the computational workload on the device. 
However, SFL introduces the following two challenges:

\textit{Challenge 1} - Low parallel efficiency due to the strong dependency between the computations carried out on the devices and server, referred to as computational locking~\cite{thapa2022splitfed,saikumar2024neuroflux},  resulting in a lower system throughput (see Section~\ref{sec2:subsec_locking}).

\textit{Challenge 2} - High communication costs are introduced due to the frequent exchange of activations and gradients between the devices and the server, thereby increasing latency and bandwidth requirements (see Section~\ref{sec2:subsec_comm}). 

We propose a novel system, \nickname that enables large models to be trained in FL without introducing the abovementioned challenges of SFL. At its core, \nickname introduces edge model overlay(s) between the device and server to facilitate the creation of a larger ensemble model that can be scaled both horizontally (model boosting) and vertically (model bagging)~\cite{arpit2022ensemble}. 
The ensemble model is created by augmenting the original (small) model trained by the FL system with any or all of the larger models trained in the overlays. 
This is referred to as \textit{`Augmented FL'} (AFL). 
Unlike SFL, AFL does not modify or replace the FL workflow and the proposed overlays are decoupled from and executed in parallel to the original FL system. 
To achieve the above, \nickname incorporates three modules: the \textit{hierarchical activation replay cache}, the \textit{convergence-aware communication controller} and the \textit{ensemble inference module}. The \textit{hierarchical activation replay cache} enables \nickname to decouple AFL from FL. The \textit{convergence-aware communication controller} reduces the AFL communication costs. Finally, the \textit{ensemble inference module} augments the original FL models at the end of training.

This work makes the following contributions:

(1) A new system, \nickname to create larger ensemble models that have up to 17.77\% higher accuracy than FL models without increasing computational overheads on the device. 

(2) A novel method, AFL to reduce computational dependencies in SFL accelerating training by up to 6.9$\times$ and reducing communication by 7.17$\times$. 

%% file: related_work.tex
\section{Related Work}
\label{sec:related_work}
Training large models in FL remains challenging due to the computational and memory constraints of participating devices. There are two categories of methods presented in the literature to reduce the computational workload and memory requirements of on-device models in FL. The first is partial training and the second is SFL. 

In \textbf{partial training} methods in FL, devices extract a sub-model from the global model using various sampling strategies, such as random selection~\cite{diao2020heterofl}, rolling selection~\cite{alam2022fedrolex}, or pruning-based selection~\cite{jiang2022fedmp}, and train only the sub-model. Despite their computational efficiency, partial training methods introduce further challenges. Since each device trains only a subset of parameters from the global model, each parameter receives fewer updates per FL round, resulting in slower convergence~\cite{cheng2022does}. Moreover, aggregating sub-models extracted from different locations further degrades the model's accuracy compared to FL training. In some cases, these methods offer no advantage over training smaller models in FL, limiting their effectiveness in scaling FL to larger models~\cite{cheng2022does}.

The \textbf{SFL} method partitions the on-device model across the device and a server unlike classic FL, where each device trains the entire model locally. Specifically, the earlier layers of the global model are trained on devices, while the later layers are offloaded to a cloud server for training~\cite{thapa2022splitfed}.
Despite its advantages, SFL suffers from low parallel efficiency due to computational locking—a strong interdependence between device and server computations. In addition, SFL incurs high communication overhead due to frequent exchanges of activations and gradients, leading to reduced system throughput, increased latency, and higher bandwidth usage.

Unlike the existing methods discussed above, the proposed system \nickname improves the learning performance of FL by creating a larger ensemble model without additional computational overhead to devices. Furthermore, the training of augmented models used for the ensemble avoids computational locking and incurs lower communication costs than those seen in SFL.

%% file: challenges.tex
\section{Computational Locking and Communication Overhead in SFL}
\label{sec:challenges}
In SFL, the later layers of the model trained on a device are offloaded to the server to alleviate the computation burden on devices~\cite{wu2022fedadapt}. However, compared to classic FL, SFL systems have strong computational dependencies and experience high communication overhead between the devices and the server.

\subsection{Computational Locking}
\label{sec2:subsec_locking}
\textbf{Model splitting in SFL.}
In SFL, the full-size model \(\bm{\Theta}\) is divided into two parts: a device-side part, \(\bm{\Theta_{[:j]}}\), and a server-side part, \(\bm{\Theta_{[j:]}}\), where \(j\) represents the split point. Specifically, layer \(j\) acts as the boundary, with \(\bm{\Theta_{[:j]}}\) containing all layers up to and including layer \(j\), and \(\bm{\Theta_{[j:]}}\) comprising all layers following layer \(j\). Therefore, the full model is represented as \(\bm{\Theta} = \{\bm{\Theta_{[:j]}}, \bm{\Theta_{[j:]}}\}\). \(\bm{\Theta_{[:j]}}\) is trained on the device, while \(\bm{\Theta_{[j:]}}\) is offloaded to and trained on the server. In contrast, classic FL trains the entire model \(\bm{\Theta}\) locally on a device.

\input{fig/fig1}

\textbf{Forward and backward locking.}
The mini-batch gradient descent algorithm is usually the method for updating parameters during model training~\cite{hinton2012neural}. Figure~\ref{fig:1} illustrates the computational dependencies in SFL and \nickname.
Given a batch of input data, devices perform forward computation on \(\bm{\Theta_{[:j]}}\). The activations generated by \(\bm{\Theta_{[:j]}}\), denoted as \(a_j\), are then transmitted to the cloud server for forward computation using \(\bm{\Theta_{[j:]}}\). The cloud server must wait for the devices to complete the forward computation using \(\bm{\Theta_{[:j]}}\) before it can execute forward computation using \(\bm{\Theta_{[j:]}}\), causing \textit{forward locking}. Similarly, after completing the forward computation and moving to the backward computation, the server first calculates the gradients \(\bm{\nabla \Theta_{[j:]}}\). The gradient of \(a_j\), denoted as \(\nabla a_j\), is then required to be sent back to devices for computing \(\bm{\nabla \Theta_{[:j]}}\). This introduces \textit{backward locking}, as devices must wait to receive \(\nabla a_j\) before calculating \(\bm{\nabla \Theta_{[:j]}}\). \textit{Forward locking} and \textit{backward locking} reduce parallel efficiency compared to classic FL, thereby decreasing throughput in SFL systems. This motivates the design of the \textit{hierarchical activation replay cache} module in \nickname to decouple computations between devices and edge layers. Consequently, \nickname eliminates forward and backward locking.

\subsection{Communication Overhead}
\label{sec2:subsec_comm}
Another challenge in SFL systems is the communication overhead, which increases latency and raises bandwidth usage.

\textbf{Communication of activation and gradient in SFL.}
Compared to classic FL, SFL introduces additional communication costs for transferring the activations (\(a_j\)) and corresponding gradients (\(\nabla a_j\)). These costs are substantial and proportional to the combined size of the datasets on devices and the number of local training iterations~\cite{singh2019detailed}. The communication costs also depend on the size of the activations and gradients, which may be larger than the original data~\cite{wu2024ecofed}.

\textbf{Communication latency and bandwidth consumption.}
Transferring the activation \(a_j\) and gradients \(\nabla a_j\) increases both communication latency and bandwidth consumption. For instance, SFL can introduce up to 800$\times$ more cumulative communication time and 610 $\times$ more bandwidth consumption compared to FL for transferring activations and gradients to the cloud (see Section~\ref{sec5:sys_per}).
This motivates the design of a \textit{convergence-aware communication controller} module in \nickname to reduce the communication overhead.

%% file: fig/fig1.tex
\begin{figure}
		\centering
		\includegraphics[width=0.5\textwidth]{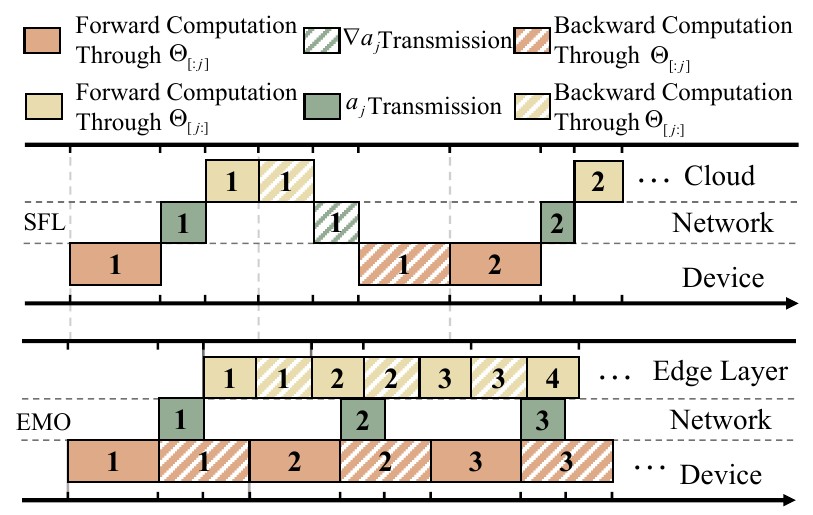}
		\caption{Computational dependencies in SFL and \nickname.}
		\label{fig:1}
         \vspace{-2mm} 
\end{figure}

%% file: elo-v2.tex
\section{\nickname: Edge Model Overlay(s)}
\label{sec:elo}

\input{fig/fig4}

\subsection{Overview}
\label{sec3:}
The \nickname architecture is shown in Figure~\ref{fig:4}. 
Augmented FL (AFL) in \nickname is hosted alongside devices on edge servers within a local area network (LAN). 
AFL runs parallel with the FL and trains larger models - the original FL model can be connected to the larger models.
The \textit{activation replay cache} module (see Section~\ref{sec3:cache}) decouples AFL from the original FL system. \nickname leverages the available activations stored in the \textit{activation replay cache}, eliminating the need to wait for new device activations (forward locking).

During AFL, activations generated by the FL model must be transferred from devices to edge servers to update the \textit{activation replay cache}. 
The transmission in \nickname occurs within a LAN. Given a LAN's higher bandwidth than a wide area network (WAN), this approach is more communication-efficient than classic SFL, which transmits activations to a cloud server via a WAN.
Additionally, the \textit{communication controller} module in \nickname further optimizes activation transfer by adjusting communication intervals (see Section~\ref{sec3:controller}).

After completing AFL, the original FL model and the larger model(s) trained in \nickname are sent to the cloud server for aggregation into an ensemble model. Both horizontal scaling (model boosting) and vertical scaling (model bagging) are employed on the cloud server to create the ensemble model (see Section~\ref{sec3:assemble}). The number of models trained in AFL depends on the number of overlays on the edge servers.

\input{fig/fig5}
\subsection{Activation Replay Cache}
\label{sec3:cache}
The first module considered is the \textit{activation replay cache}, which decouples AFL from FL. 
\nickname eliminates forward locking seen in SFL by reusing activations received from previous communications. Classic SFL discards activations after a single use. 
\nickname preserves the original FL training workflow, thus avoiding backward locking.
Two methods underpin the activation replay cache: hierarchical activation caching and activation replay with cache sampling.

\textbf{Hierarchical activation caching.}
The challenge in building an activation cache using previous communications is to store activations on the edge server efficiently.
The size of the activation cache can be substantial, as it depends on the number of activations from all devices, which is determined by both the data size and the number of local training iterations.
\nickname employs a hierarchical caching architecture to mitigate the potentially large storage requirements. 
As shown in Figure~\ref{fig:3}, \nickname indexes activations by device ID and batch ID.
For each index (a device-batch ID pair), the most recently updated activation is stored in the h5 file format\footnote{An h5 file is a file format used to store large amounts of numerical data.}
Additionally, \nickname builds a second-level activation cache on-disk containing all activations while maintaining a first-level GPU memory cache generated by a sampling policy.
This hierarchical caching method allows \nickname to store all activations on large disks while minimizing the I/O overhead associated with transferring activations from disk to GPU memory.

\textbf{Activation replay with cache sampling.}
The second key consideration in the \textit{activation replay cache} module is to update the first-level GPU cache.  
In \nickname, the update involves sampling activations from the disk cache to the GPU cache.
Specifically, after each training iteration using activations stored in the GPU cache, the cache is refreshed by sampling new activations from the disk cache.
\nickname employs a mixed sampling policy: newly received activations not loaded into the GPU cache are directly selected for training, ensuring the prioritization of fresh data. A random sampling strategy is used if no new activations are available (i.e., all activations have previously been loaded into the GPU cache).
This policy prioritizes new activations and switches to uniform sampling once all have been utilized. 
The random sampling policy can be replaced with importance sampling, where activations are selected based on their associated training loss~\cite{katharopoulos2018not}.

\subsection{Convergence-aware Communication Controller }
\label{sec3:controller}
To reduce the communication overhead, \nickname incorporates a \textit{convergence-aware communication controller} to monitor activation convergence and dynamically adjust the communication frequency based on the state of convergence of the activations.

\textbf{Activation convergence.}
Recent studies have demonstrated that, during model training, the earlier layers converge faster than the later layers — a phenomenon referred to as the bottom-up learning dynamic~\cite{raghu2017svcca}.
This observation provides an opportunity for \nickname to reduce activation transfer costs by limiting the transfer of activations from the original FL model as it gradually converges.
During the early stages of training, when the distribution of activations undergoes significant changes, \nickname collects activations from devices more frequently and updates the disk cache accordingly.
As training progresses and the distribution of activations stabilizes, the intervals for activation collection and cache updates are gradually extended. During these intervals, \nickname reuses the existing cache, reducing communication overhead.
To assess activation convergence, \nickname employs Singular Vector Canonical Correlation Analysis (SVCCA)~\cite{raghu2017svcca}. The SVCCA score is a normalized metric ranging from 0 to 1 to quantify the correlation between two activations; a higher score indicates higher convergence~\cite{raghu2017svcca}.

\input{algorithm1}

\textbf{Convergence-aware activation transmission protocol.}
Algorithm~\ref{algorithm1} outlines the activation transmission protocol implemented in the \textit{communication controller}, which dynamically adjusts activation transmission intervals based on the convergence of activations.
The decision flow for transferring the activation batch \((i,j)\) is as follows: the controller first checks whether the batch ID is already stored in the ID cache \(\mathbb{B}\). If the batch ID is not cached, the device must transmit this batch. Next, the controller checks the interval between the last activation transmission and the current attempt. The device sends activations \(a_{(i,j)}\) only if the predefined transmission interval has been reached; otherwise, the activations are not sent to \nickname, and the communication interval counter is incremented. The communication interval, ID cache, and counter are logged in a log file for querying.
Upon receiving the activation, \nickname checks whether the batch is already cached. If it is, \nickname retrieves the previously stored activation \(a_{(i,j)}^c\) and calculates the SVCCA score \(s_{(i,j)}\) between the new and cached activations. The cached activation is then updated with the new one, and the communication interval is adjusted using \( i_{(i,j)} = \frac{1}{1 - s_{(i,j)}} \). 
This convergence-aware interval adaptation mechanism effectively minimizes redundant communication of similar activations while ensuring that non-trivial activation updates are transmitted.

\subsection{Ensemble inference}
\label{sec3:assemble}
After both FL and AFL are completed, the original FL model, denoted as \(M_{FL}\), along with the models trained in the overlay \(M_{EMO}^i\) generated on edge server \(i\) by \nickname, are sent to the cloud for aggregation into an ensemble model.
AFL can be executed across \(N\) edge servers, resulting in the generation of multiple models within \nickname.

Two categories of model aggregation are employed. The first is \textit{horizontal aggregation}, where the original FL model \(M_{FL}\) is horizontally connected to each overlay model \(M_{EMO}^i\), also referred to as model boosting~\cite{arpit2022ensemble}. This results in a deeper model \((M_{FL}, M_{EMO}^i)\), increasing the number of layers in the original FL model \(M_{FL}\). 
The second is \textit{vertical aggregation}, where the final outputs (logits) from the \(N\) pairs of original FL models and their corresponding overlay models \((M_{FL}, M_{EMO}^i)\) are averaged before being used for class prediction. This approach is referred to as vertical aggregation or model bagging~\cite{arpit2022ensemble}.
The final ensemble model is deployed in the cloud for inference.

%% file: fig/fig4.tex
\begin{figure}
		\centering
		\includegraphics[width=0.5\textwidth]{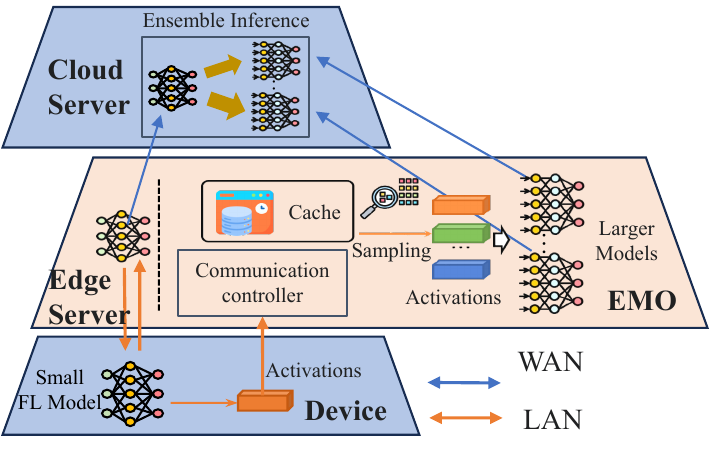}
		\caption{The \nickname architecture.}
		\label{fig:2}
         \vspace{-2mm} 
\end{figure}

%% file: fig/fig5.tex
\begin{figure*}
		\centering
		\includegraphics[width=0.96\textwidth]{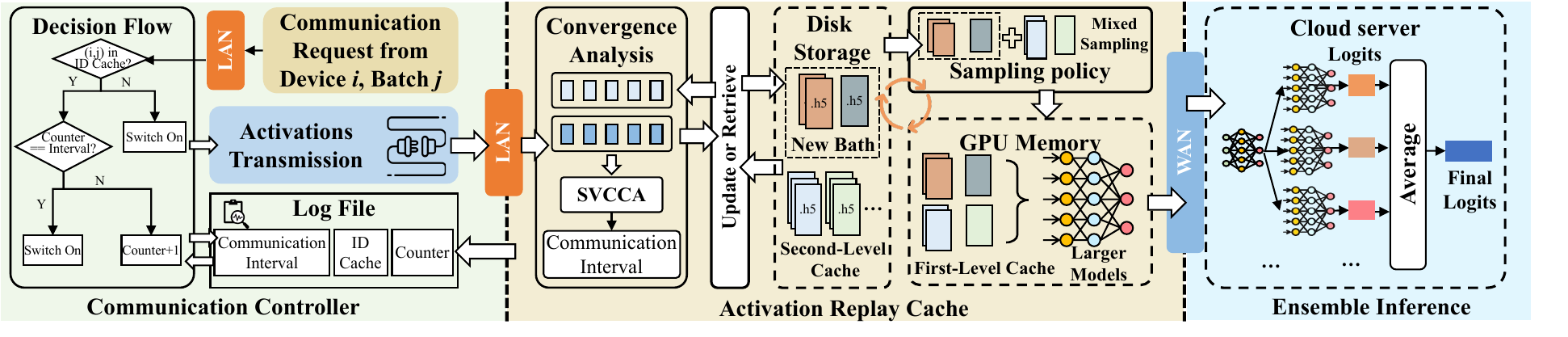}
		\caption{The activation replay cache, convergence-aware communication controller and ensemble inference modules in \nickname.}
		\label{fig:3}
        \vspace{-1mm} 
\end{figure*}

%% file: algorithm1.tex
\begin{algorithm}[t]
\caption{Convergence-aware activation transmission protocol}
\label{algorithm1}
\textbf{Set up:} Receive the transfer request for activation batch \((i, j)\) from device \(i\) for batch \(j\). The current log file consists of an ID cache \(\mathbb{B}\), the communication interval for this batch \(i_{(i,j)}\), and the interval counter \(c_{(i,j)}\).\\
\If{$(i,j) \notin \mathbb{B}$}{
    Require the activation batch \(a_{(i,j)}\) to be sent;\\
    Initialize $s_{(i,j)}=0$, $c_{(i,j)}=0$. 
}
\Else{
    \If{$c_{(i,j)} == i_{(i,j)}$ }{
    Require the activation batch \(a_{(i,j)}\) to be sent.\\
    Extract activation $a_{(i,j)}^c$ from the disk cache;\\
    Compute the SVCCA~\cite{raghu2017svcca} score of $a_{(i,j)}$ by  \\
    $s_{(i,j)}\leftarrow SVCCA(a_{(i,j)}^c,a_{(i,j)})$;\\
    Replace the activation in disk cache $a_{(i,j)}^c\leftarrow a_{(i,j)}$;\\
    Update the communication interval $i_{(i,j)}\leftarrow \frac{1}{1-s_{(i,j)}}$;\\
    Reset interval counter $c_{(i,j)}=0$.
    }        
    \Else{
    Skip sending this batch activation $a_{(i,j)}$.
    $c_{(i,j)}\leftarrow c_{(i,j)}+1$;
    }
}
\end{algorithm}

%% file: evaluation-v2.tex
\section{Evaluation}
\label{sec:evaluation}
We evaluated \nickname against two baselines to assess both the learning and system performance using metrics such as accuracy, communication cost, and training time.

\subsection{Experimental Setup}

\textbf{Testbed.}
We developed a three-tier system comprising end devices, two edge servers, and a cloud server. The end devices are organized into two clusters: Raspberry Pi 4 Model B single-board computers and NVIDIA Jetson NX devices equipped with GPUs. The edge servers are powered by NVIDIA RTX 3070 Ti GPUs, while the cloud server operates on an NVIDIA P100 GPU hosted on Alibaba Cloud.
All end devices and edge servers are interconnected via a LAN with symmetrical upload/download bandwidths of 800 Mbps. Additionally, both the devices and edge servers are connected to the cloud server over a WAN, with an upload/download bandwidth of 100 Mbps.

\textbf{Baselines.}
We evaluated \nickname against two baselines: \textbf{FL:} The classic FL approach is applied in two configurations: \textbf{FL-Small}, where a small model is trained on devices, and \textbf{FL-Large}, where a large model is fully trained on devices. In both cases, the entire model resides on devices during training. \textbf{SFL:} In the SFL baseline, the top seven layers of the large model are offloaded and trained on a server. We implement two SFL variants: \textbf{SFL-Cloud}, where the later layers of the model are offloaded to a cloud server, and \textbf{SFL-Edge}, where they are offloaded to an edge server. \textbf{\nickname:} For \nickname, we implement two edge model overlays on separate edge servers, covering 40\% and 60\% of the devices, respectively.

\input{exp/tab1}
\input{exp/tab2}

\textbf{Training Setup.}
The training tasks are conducted on the CIFAR-10~\cite{krizhevsky2009learning} and CIFAR-100~\cite{krizhevsky2009learning} datasets, using VGG and MobileNet~\cite{howard2017mobilenets} models, respectively. The datasets are partitioned across devices in a non-independent and non-identically distributed (non-IID) manner using the Dirichlet distribution method~\cite{hsu2019measuring}. 
In each training round, 20 devices are randomly selected from a total of 100 devices. Training is carried out over 200 rounds with a learning rate of 0.01. The batch size is set to 16, and the aggregation method used for both FL and SFL is FedAvg~\cite{mcmahan2017communication}. 
Table~\ref{table:model_partition} summarizes the model configurations employed for different methods and datasets.

\subsection{Learning Performance}
\input{exp/exp1}

Figure~\ref{fig:4} shows the test accuracy curves for the baselines and \nickname. On CIFAR-10, FL-Large, SFL, and \nickname achieve higher accuracy in the early stages of training. However, on CIFAR-100, SFL and \nickname significantly outperform the FL methods during initial training. In the later stages, \nickname consistently surpasses all baselines on both CIFAR-10 and CIFAR-100.
Table~\ref{table:exp1} reports the highest test accuracy achieved by \nickname and the baseline methods. On CIFAR-10, FL-Large achieves accuracy comparable to FL-Small. However, on CIFAR-100, FL-Large outperforms FL-Small by 8.71\%, demonstrating the benefits of increased model size in FL training. Across both datasets, \nickname outperforms all baselines by a considerable margin. Specifically, it achieves up to 3.29\% and 17.77\% higher accuracy than the FL methods on CIFAR-10 and CIFAR-100, respectively. Compared to SFL, \nickname improves accuracy by 2.38\% on CIFAR-10 and 1.38\% on CIFAR-100.

\subsection{System Performance}
\label{sec5:sys_per}
For system performance, we report the average communication cost and training time per round, where a round is defined as a complete cycle of local training on the devices followed by aggregation on the server.

\textbf{Communication cost.}
The communication cost of \nickname and the four baselines for a single training round are presented in Table~\ref{table:comm}. On CIFAR-10, the size of the VGG model is the main contributor to communication overhead, requiring 0.6 GB for FL-Small and 2.91 GB for FL-Large. SFL introduces an additional 0.3 GB of communication overhead compared to FL-Small, totaling 0.9 GB, while \nickname reduces this to 0.83 GB.
Using CIFAR-100 and given that the MobileNet model is lightweight, SFL results in the highest communication cost at 2.44 GB due to the communication of activations and gradients. In contrast, \nickname achieves a substantial reduction in communication overhead, lowering it to 0.34 GB, which is a 7.17$\times$ decrease compared to SFL.

\input{exp/tab3}
\input{exp/exp2}

\textbf{Training time.}
The average training time of a round and communication time for \nickname and the baselines are shown in Figure~\ref{figure:5}. Using both CIFAR-10 and CIFAR-100, \nickname achieves the lowest training time.
For CIFAR-10, communication time is the dominant factor for FL-Small, FL-Large, and SFL-Cloud. However, SFL-Edge and \nickname benefit from higher bandwidth between devices and edge servers, resulting in reduced communication time. 
On CIFAR-100, communication time is not the bottleneck for FL-Small and FL-Large due to the smaller size of the MobileNet model. Nevertheless, both SFL-Cloud and SFL-Edge introduce additional communication time as a result of transferring activations and gradients to remote servers.
Overall, compared to FL and SFL, \nickname reduces round training time by up to 12.5$\times$ and 6.9$\times$, respectively.

%% file: exp/tab1.tex
\begin{table}
	\centering
	\caption{Models used for evaluating different methods. The number following each model represents the total number of layers. \textbf{C} denotes convolutional layers and \textbf{F} represents fully connected layers (both are preceded by the number of layers).}
    \begin{tabular}{ m{1cm}<{\centering}m{1.2cm}<{\centering}m{1.2cm}<{\centering}m{1.6cm}<{\centering}m{1.6cm}<{\centering} }
    \Xhline{2\arrayrulewidth}
    \multirow{2}{*}{\textbf{Methods}}&
    \multicolumn{2}{c}{\textbf{CIFAR-10}}&
    \multicolumn{2}{c}{\textbf{CIFAR-100}}\\
    \cline{2-5}
    &\textbf{Device}&\textbf{Server}&\textbf{Device}&\textbf{Server}\\
    \hline
    FL-Small
    & 
    VGG-6
    (4C + 2F)&
    N/A &
    MobileNet-6 
    (5C + 1F)&
    N/A \\
    \hline

    FL-Large
    & 
    VGG-11 
    (8C + 3F)&
    N/A &
    MobileNet-12 
    (11C + 1F)&
    N/A \\
    \hline

    SFL
    & 
    VGG-4 
    (4C) 
    &
    VGG-7 
    (4C + 3F)&
    MobileNet-5
    (5C)&
    MobileNet-7 
    (6C + 1F)\\
    \hline
    
    \textbf{\nickname}
    & 
    VGG-6 
    (4C + 2F)&
    VGG-7
    (4C + 3F)&
    MobileNet-6 
    (5C + 1F) &
    MobileNet-7 
    (6C + 1F) \\
    
    \Xhline{2\arrayrulewidth}
    \end{tabular}
    \label{table:model_partition}
\end{table}

%% file: exp/tab2.tex
\begin{table}
	\centering
	\caption{Highest test accuracy of \nickname and baselines.}
    \begin{tabular}{ P{1.5cm} P{1.2cm} P{1.2cm} P{1.2cm} P{1.2cm} }
    \Xhline{2\arrayrulewidth}
    \multirow{2}{*}{\textbf{Dataset}}&
    \multicolumn{4}{c}{\textbf{Methods}}\\
    \cline{2-5}
    &\textbf{FL-Small}&\textbf{FL-Large}&\textbf{SFL}&\textbf{\nickname}\\
    \hline
    CIFAR-10
    & 
    76.57\% &
    76.22\% &
    77.13\% &
    \textbf{79.51\%} \\
    \hline

    CIFAR-100
    & 
    24.92\% &
    33.63\% &
    41.31\% &
    \textbf{42.69\%} \\
    
    \Xhline{2\arrayrulewidth}
    \end{tabular}
    \label{table:exp1}
\end{table}

%% file: exp/exp1.tex
\begin{figure}[tp]
        \begin{subfigure}[b]{0.232\textwidth}
         \centering
         \includegraphics[width=\textwidth]{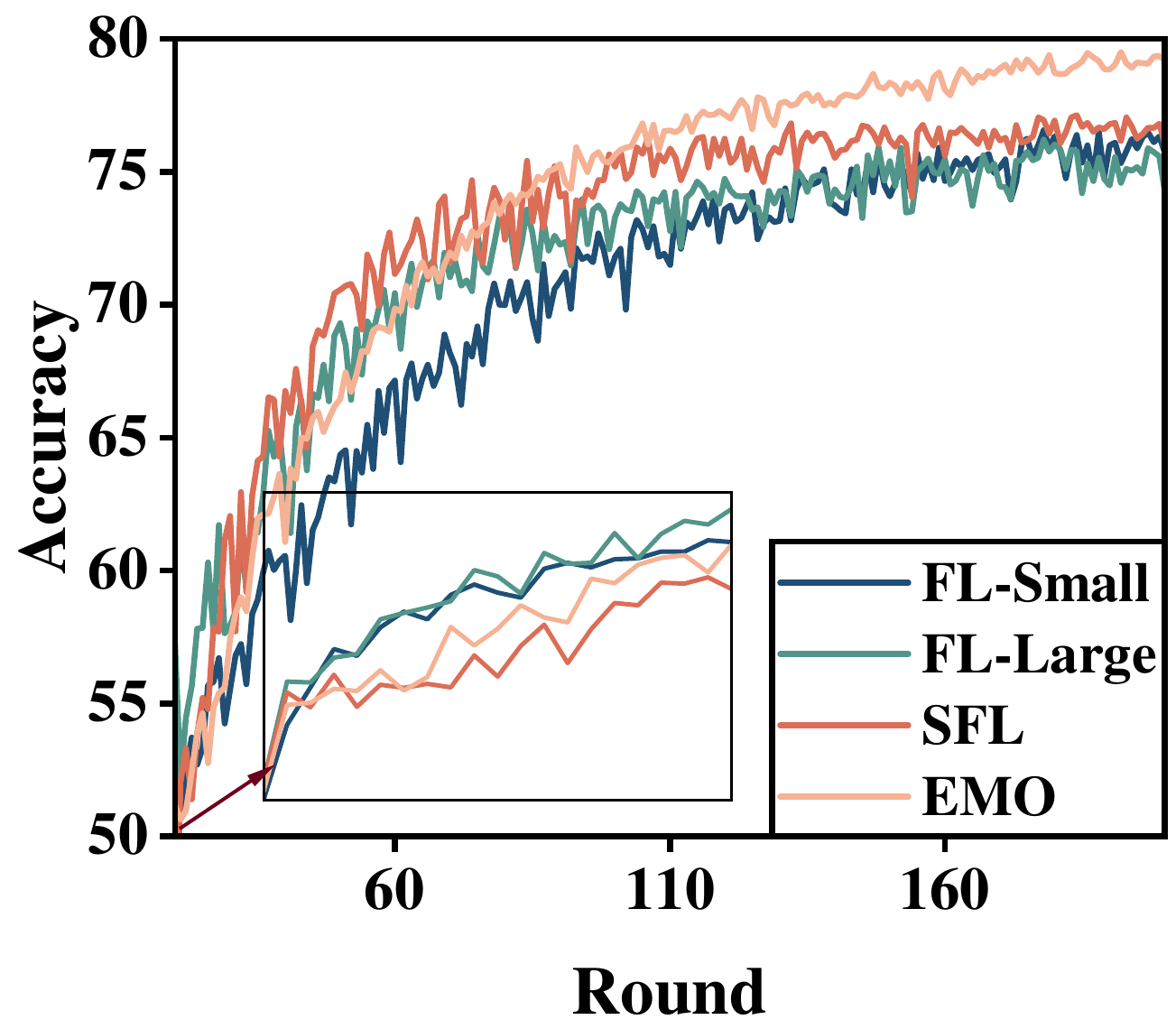}
         \caption{CIFAR-10}
         \label{fig4:cifar10}
     \end{subfigure}
     \begin{subfigure}[b]{0.24\textwidth}
         \centering
         \includegraphics[width=\textwidth]{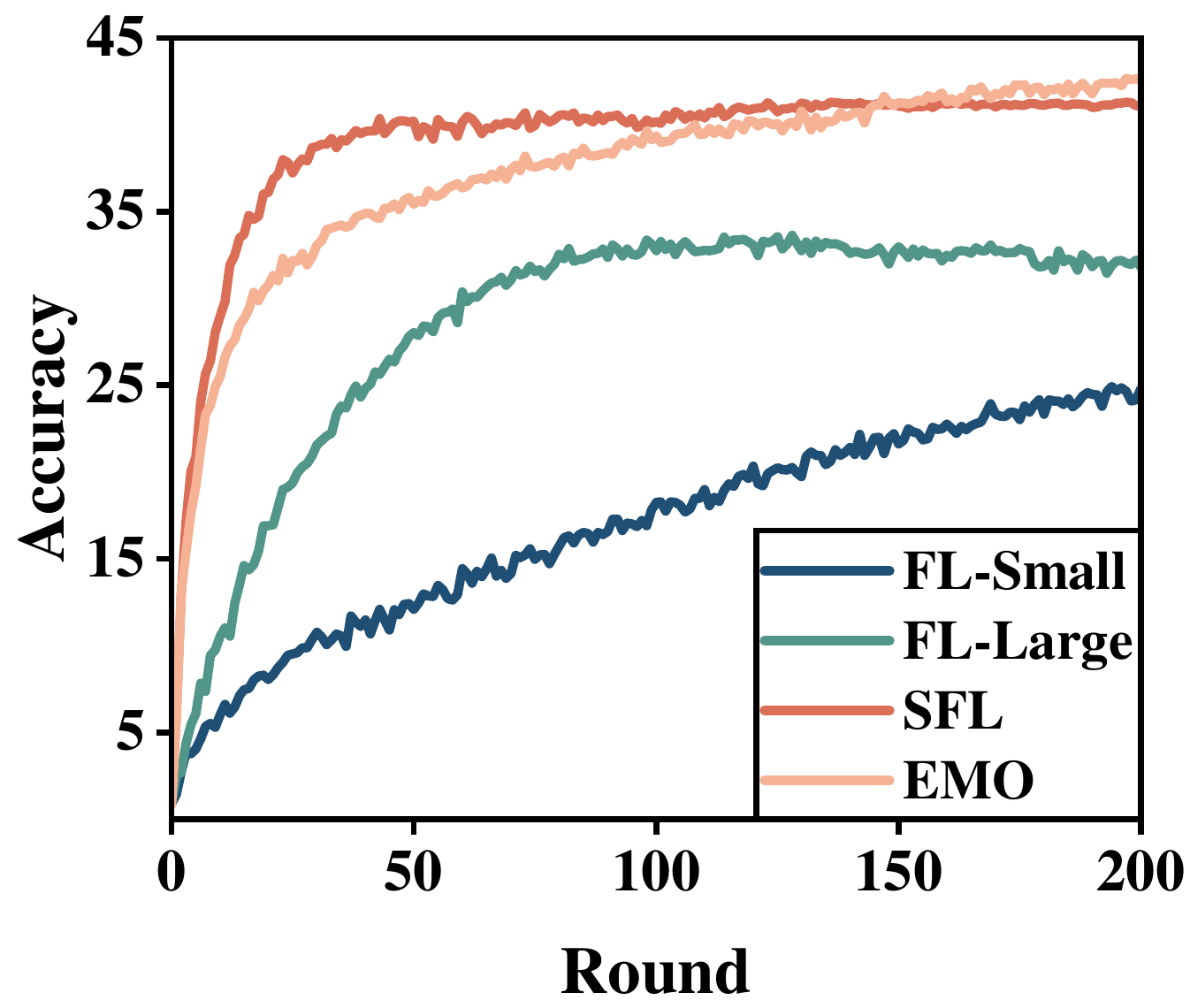}
         \caption{CIFAR-100}
         \label{fig4:cifar100}
     \end{subfigure}
		\caption{Test accuracy curves of \nickname and baselines for CIFAR-10 and CIFAR-100 datasets.}
		\label{fig:4}
        \vspace{-1mm}
\end{figure}

%% file: exp/tab3.tex
\begin{table}
	\centering
	\caption{Average communication cost per round.}
    \begin{tabular}{ P{3cm} P{2cm} P{2cm} }
    \Xhline{2\arrayrulewidth}
    \multirow{2}{*}{\textbf{Methods}}& 
    \multicolumn{2}{c}{\textbf{Communication cost}}\\
    \cline{2-3}

     &  CIFAR-10&CIFAR-100\\
     \hline
     FL-Small&  0.60 GB&  0.004 GB\\
     \hline
     FL-Large&  2.91 GB&  0.12 GB\\
     \hline
     SFL&  0.90 GB&  2.44 GB\\
     
     \hline
     \textbf{\nickname}&  \textbf{0.83 GB}&  \textbf{0.34 GB}\\

     \Xhline{2\arrayrulewidth}
    \end{tabular}
	\label{table:comm}
\end{table}

%% file: exp/exp2.tex
\begin{figure}[tp]
    \centering
    \begin{subfigure}[b]{0.3\textwidth}
         \centering
         \includegraphics[width=\textwidth]{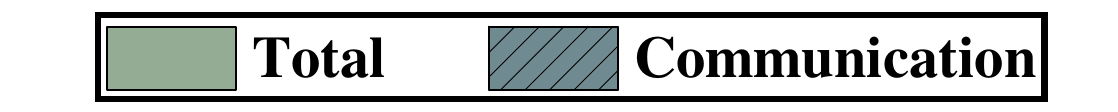}
     \end{subfigure} 
    \begin{subfigure}[b]{0.235\textwidth}
         \centering
         \includegraphics[width=\textwidth]{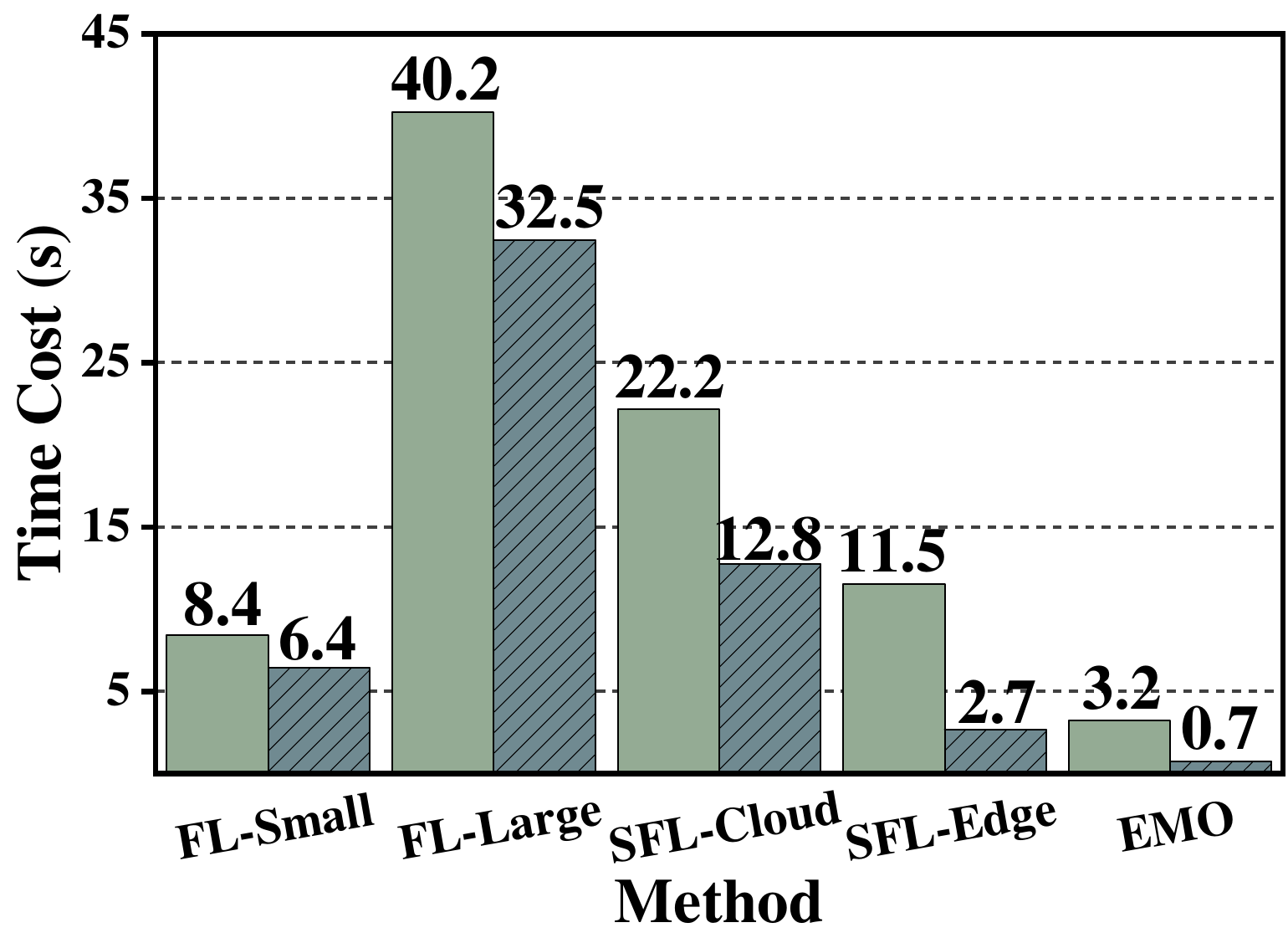}
         \caption{CIFAR-10.}
         \label{fig5:}
     \end{subfigure}
     \begin{subfigure}[b]{0.24\textwidth}
         \centering
         \includegraphics[width=\textwidth]{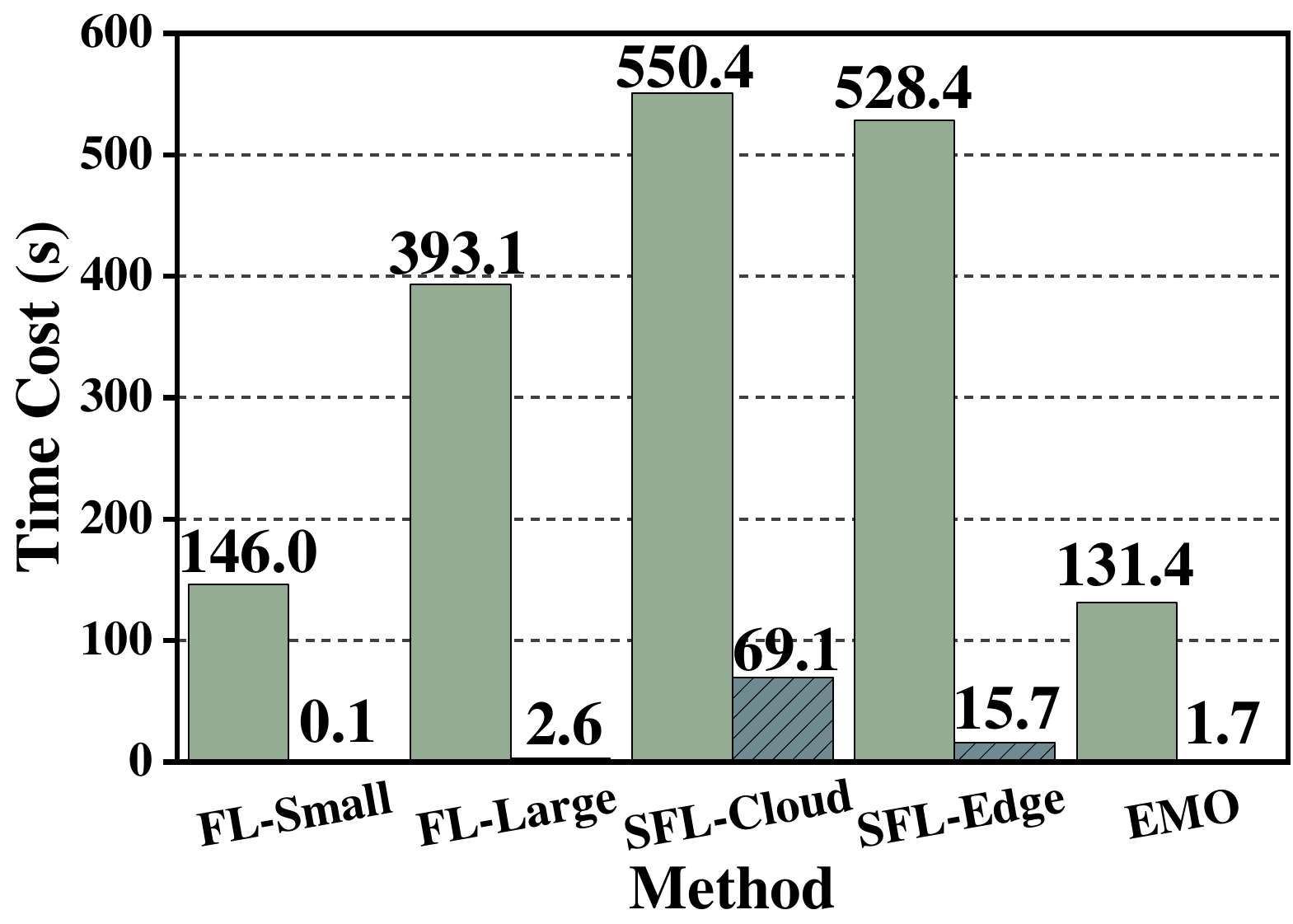}
         \caption{CIFAR-100.}
         \label{fig5:}
     \end{subfigure}
		\caption{Average training time per round for different methods, including its communication and total time.}
		\label{figure:5}
        \vspace{-2mm}
\end{figure}

%% file: conclusion.tex
\section{Conclusion and Future Work}
\label{sec:conclusion}
We propose \nickname, Edge Model Overlay(s), a system that augments FL by scaling the size of trained models while addressing the challenges associated with SFL.
Unlike SFL methods, \nickname is decoupled from the original FL system and is executed in parallel without modifying the FL workflow. The overlays train larger models on edge servers that can be connected to the smaller FL models trained on devices. 
Our experiments demonstrate that \nickname improves FL training accuracy with augmented models by up to 17.77\% compared to FL training with a small model, while also reducing training time. Additionally, \nickname decreases communication overhead by up to 7.17$\times$ and cuts training time by 6.9$\times$ compared to SFL.
In future work, we will investigate techniques to enhance privacy of the activations cached in \nickname.